\documentclass{article}

\usepackage{arxiv}
\usepackage[utf8]{inputenc}
\usepackage[T1]{fontenc}
\usepackage{hyperref}
\usepackage{url}
\usepackage{booktabs}
\usepackage{amsfonts}
\usepackage{amsmath}
\usepackage{graphicx}
\usepackage{tabularx}
\usepackage{longtable}
\usepackage{float}
\usepackage{microtype}
\usepackage{doi}

\title{Can Urban Blight Be Accessed with Vision-language Models: A Case Study in Detroit}

\author{
Xiaohao Yang\thanks{Corresponding author: \texttt{xiaohaoy@umich.edu}} \\
School for Environment and Sustainability; School of Information\\
University of Michigan, Ann Arbor, USA\\
\And
Aohua Tian \\
School for Environment and Sustainability\\
University of Michigan, Ann Arbor, USA\\
\And
Derek Van Berkel \\
School for Environment and Sustainability\\
University of Michigan, Ann Arbor, USA\\
\And
Xu Qiang \\
School for Environment and Sustainability\\
University of Michigan, Ann Arbor, USA\\
\And
Mark Lindquist \\
School for Environment and Sustainability\\
University of Michigan, Ann Arbor, USA
}

\renewcommand{\shorttitle}{Can Urban Blight Be Accessed with Vision-language Models?}

\hypersetup{
pdftitle={Identifying and Mapping Urban Blight in Detroit Using Vision-language Models},
pdfauthor={Xiaohao Yang, Derek Van Berkel, Aohua Tian, Xu Qiang, Mark Lindquist},
pdfkeywords={Urban blight, Residential housing condition, Vision-language models, Street view images, Ensemble learning},
}

\begin{document}
\maketitle

\begin{abstract}
Addressing urban blight has seen increased focus in the past 15 years. Assessing urban blight is essential for guiding urban planning, targeting rehabilitation, and safeguarding public health, yet traditional residential blight surveys are difficult to maintain at scale due to the labor-intensive cost and long-term cycle. This study introduced a scalable framework for estimating residential blight using open-source large vision-language models on multiple views. Structured prompts guided models to evaluate housing attributes, including roof integrity, wall damage, and broken or boarded openings, producing both binary assessments and probabilistic estimates of disrepair. To evaluate the performance of these visual assessments, we compared professional human annotations of these features across several models, including an ensemble stacking approach based on XGBoost and a weighted scoring system. Results showed that (i) multiple street views can contribute to the improvement of accuracy, (ii) large vision-language models have different strengths of inference, (iii) the ensemble learner outperforms individual base models, enhancing robustness across all residential conditions and blight assessment. The practical application of the method allows low-cost tracking and managing of housing stock conditions, providing a regularly updatable complement to traditional blight surveys.
\end{abstract}


\section*{Highlights}
\begin{itemize}
\item Multiple street views can contribute to the improvement of prediction accuracy
\item VLMs have different inference strengths in detecting residential damage
\item Ensemble learners outperform individual base models
\item Low-cost method to track and update housing conditions beyond on-site surveys
\end{itemize}

\section{Introduction}
City planners are increasingly confronted with complex urban management and planning challenges, driven by the need to adapt to more frequent extreme weather events and shifting social and economic conditions such as urban blight, gentrification, and declining livability (Pinto et al., 2021; Atkinson, 2004; Jiang \& Sun, 2024). Emerging data sources and smart technologies present new opportunities to improve planning, enhance public services, boost efficiency, and build more sustainable and livable cities (Bibri, 2019). Among these efforts, accurate and scalable assessment of housing stock is critical for understanding population dynamics, identifying maintenance needs, and guiding targeted interventions (Neidert et al., 2025; U.S. Congress, 2021). Comprehensive information on habitability, yard upkeep, roof integrity, and structural disrepair--key indicators of blight--offers valuable insight into community well-being and informs decisions on code enforcement, redevelopment, and public investment. Yet, despite its importance, a comprehensive evaluation of these residential conditions remains a complex and resource-intensive task for many municipalities.

Assessment of urban blight, defined as 'deteriorating property conditions that have deleterious effects on the community in which the property is situated' (Beers et al., 2011), has become increasingly important in post-industrial cities. Decades of population loss, disinvestment, and neglect have produced long-term vacancy and housing disrepair, leaving thousands of properties deteriorated and often uninhabitable, and a challenge for city managers to tackle  (Pinto et al., 2023). Internal and external housing conditions can affect the physical and mental health of residents (Bonnefoy, 2007; Pevalin et al., 2017), while damaged building structures (e.g., leaky roofs and broken windows) can create damp and cold dwelling environments, promoting mold growth and associated with increased risks of recurrent headaches, sore throat, and respiratory diseases (Board on Health Promotion, Disease Prevention, \& Committee on Damp Indoor Spaces, 2004). In addition, degraded roofs and windows are associated with negative mental health effects (Ochodo et al., 2014).

Conventional housing assessments are usually conducted via in-situ evaluations by trained personnel using structured surveys (Hillier et al., 2003; Konomi et al., 2019; Kumagai et al., 2016; Yin and Silverman, 2015), and many cities have used these methods to conduct citywide surveys to map residential areas. While these methods can provide high-quality insights, they are time-consuming and subject to variability across assessors, leading to challenges in maintaining and updating comprehensive documentation over time. Therefore, there is a growing need for reliable, scalable, and sustainable methods that can systematically assess housing stock and residential environments to support data-driven decision-making in urban planning and policy.

To overcome these challenges, less labor-intensive methods have been developed using statistical survey data, spatial data, and imagery data to derive large-scale housing information. Aggregated-scale housing-related datasets, such as the American Housing Survey (Dewar et al., 2012; Mallach, 2018), United States Postal Service administrative data (Silverman et al., 2013), and nighttime light imagery, have been used to identify abandoned buildings. Compared to these datasets, open-access image data sources such as aerial and street view images (SVIs) can provide an individual-property-level perspective for urban house studies at the finest scale. Previously, Zou and Wang (2021) used SVIs to detect individual vacant houses. Recently, Zou and Wang (2022) proposed a framework that integrated top-down views (satellite images) and SVIs to improve the detection of housing vacancy. Emerging approaches using predictive modeling have shown promise in identifying areas of concern, but they can also inherit biases from the underlying data and lack the granularity of on-the-ground observations (Liang et al., 2024). In response, some city managers have begun exploring the use of urban street-view imagery to assess housing stock (Detroit Land Bank Authority, 2025). These efforts typically rely on systematic visual inspections of street-level and oblique photographs, guided by detailed rubrics developed by city departments. Although these image-based evaluations are critical for comprehensive urban planning and code enforcement, they remain labor-intensive and are often constrained by staff capacity.

While researchers have developed frameworks to detect abandoned and vacant houses at both regional and property-specific scales, methods for estimating housing blight remain underexplored due to the complexity of defining and assessing housing conditions. Newer methods using artificial intelligence are promising; however, pre-trained machine learning (ML) models from specific cities often struggle to generalize to new locations due to limited training samples. This is primarily because most models are not equipped to handle variations in housing appearance that were not represented during the model training stage (Zuo \& Wang et al., 2022). Moreover, unlike vacancy detection, which focuses on a single attribute, blight assessment should involve multi-attribute detections, including broken or missing windows, doors, walls, and roofs, which may not be suitable for single-attribute computer vision models to handle.

For reasoning built environment conditions, vision-language models (VLMs) may provide potential alternatives to ML models that rely on established ML ecosystems (e.g., TensorFlow and PyTorch) and require training on labeled large imagery datasets. For instance, Malekzadeh et al. (2025) found a strong alignment between human ratings and zero-shot inferences of ChatGPT for evaluating urban attractiveness using street views. Meanwhile, a single VLM can handle multi-attribute predictions. For example, Liang et al. (2025) generated building information (e.g., building type, number of floors, age, and surface material) using a fine-tuned VLM and street view images. Therefore, it is worth investigating the usage of VLMs in inferring multiple attributes of residential condition and expanding the scope of urban spatial reasoning to urban blight at an individual house level. However, the performance of open-source large VLMs in studying the residential environment has not been investigated.

This research investigates the capacity of open-source VLMs for detecting visible signs of disrepair across the city's residential building stock at the citywide scale. We introduced a scalable framework for mapping residential blight in Detroit by leveraging street view images and open-source large VLMs. Moreover, an ensemble approach, which leverages outputs from all base VLMs as inputs to achieve higher accuracy. was deployed to improve accuracy. Through results, we tried to answer four research questions: (i) Whether multiple views can contribute to the detection of housing disrepairs by open-source VLMs? (ii) What type of housing disrepairs can be recognized better than others by VLMs? (iii) To what extent does the VLM-based assessment of residential blight differ from the assessment based on human observation? (iv) Whether the accuracy of the assessment can be improved by integrating predictions by different VLMs?

\section{Related studies}
\subsection{Advancement of large language models in geospatial research}
The emergence of large language models (LLMs), and specifically VLMs, has opened new avenues for gaining spatial context and reasoning through multimodal data integration. VLMs pair computer vision with the natural-language capabilities of LLMs to provide contextual understanding of images; they can describe--and, when prompted, explain--what they depict (e.g., whether a roof is intact or a window is missing). By and large, spatial context is obtained by leveraging location-specific imagery (e.g., geotagged social media, satellite, and street-view images). Relatedly--though not confined to VLMs or image-based tasks-- geospatial research is incorporating spatial cognition into LLMs by including explicit cues (coordinates, place IDs, topological relations). Thus far, classification tasks using vision--language pipelines on multimodal spatial imagery have primarily targeted biodiversity assessment, infrastructure, and urban feature extraction (e.g., roads, power lines, rooftops), and in the context of disaster management. As an example, Gillespie et al. mapped fine-scale shifts in plant species using remotely sensed imagery and crowd-sourced iNaturalist photos (Gillespie et al., 2024). Wang et al. (2023) mapped electrical distribution grids using machine-learning models and street view images, road networks, and building maps. In the disaster-management context, CrisisMMD combines tweet text with location coordinates to detect damage and analyze response patterns during hurricanes (Alam et al., 2018).

Urban-specific studies have been a main focus in this context. For example, Liang et al. (2025) demonstrated that fine-tuned VLMs can infer building-level properties (structure type, floors, construction age, material condition) directly from Mapillary street-view imagery, while Opencity3D (Bieri et al., 2025) uses VLM to reconstruct buildings based on remote sensing. UrbanCLIP (Yan et al., 2024) improved urban region profiling by integrating text into image representations via contrastive language--image pretraining. Despite these advances, relatively few studies have directly applied VLMs to the systematic assessment of physical housing conditions.

LLMs with spatial cognition (awareness of spatial relationships) have mostly been applied to environmental perception tasks. CityGPT (Feng et al., 2024) and UrbanGPT (Li et al., 2024), for example, are models specifically designed for urban spatial understanding by integrating spatio-temporal dependencies into LLM architectures. While these models represent critical progress toward complex urban reasoning, their primary focus remains on simulating urban dynamics rather than assessing the physical conditions of built environments. Parallel research has explored the use of generative LLM agents for subjective perception collection. Verma et al. (2023) employed LLMs to simulate human emotional responses to streetscapes, offering a scalable alternative to traditional perception surveys.

\subsection{Expanding geospatial analysis through ensemble learning}
Comparative evaluations of LLMs and VLMs, such as that by Malekzadeh et al. (2025), which contrasted AI-generated and human assessments of urban attractiveness, have highlighted both the capabilities and limitations of current LLM-based perception models, particularly in capturing nuanced contextual information. Open questions remain about the trustworthiness of LLMs and VLMs, and there are growing calls to score and report consistency and robustness when deploying language models for real-world urban decision-making (Dona et al., 2024).  Bias or inaccuracies in LLMs often stem from the out-of-the-box capability (from zero-shot and few-shot inference), differences in architecture, parameter count, tokenization, and training methods and datasets (e.g., under-/over-representation, label noise, subjective annotations, and domain shift), which can systematically affect predictions across groups, places, or contexts (Chen et al., 2025). Variation across models highlights the need to account for differences in accuracy by subgroup and geography (e.g., reporting disaggregated metrics, calibrating thresholds, and tailoring deployment). However, practical guidance on how to operationalize these safeguards in applied settings remains limited. While Wolpert \& Macready (2002) claimed that no algorithm is superior to all alternatives under every condition, ensemble learning is one of the machine learning approaches to address this situation.

Ensemble stacking, also known as stacked generalization, may help address this accuracy challenge by combining base-model predictions. The approach combines predictions of base models to train a higher-level model (meta-learner) on the same dataset (Wolpert et al., 1992). In general, a meta-learner learns how to weight the predictions of different base models and often performs better than an individual base model (Ting \& Witten, 1999; Cho et al., 2020; Lv et al., 2022). Ensemble learning has been widely used in remote sensing research (Zhang et al., 2022), applying random forest and XGBoost (eXtreme Gradient Boosting) as ensemble learners in the prediction of yield (Cao et al., 2022), forest biomass and change (Ghosh et al., 2021; Healey et al., 2018), and natural hazards (Zeng et al., 2023). Moreover, XGBoost has been adapted for ensemble stacking learning in urban spatial research. For instance, Zhou et al. (2020) classified land use using an XGBoost learner trained on multiple features extracted from imagery and social sensing data.

These examples indidate it may be more advantageous to evaluate multiple off-the-shelf VLMs in parallel and combine their outputs rather than relying on a single model chosen by public rankings. While the ensemble learning approach has been introduced to the application of integrating multiple LLMs to harness their distinct strengths (Niimi, 2025; Chen et al., 2025), image-based geospatial tasks rarely incorporate this method with the inference of multiple VLMs, especially for residential conditions.

\section{Method}
Our research consists of three parts: inference of building conditions using VLM and street view imagery, blight estimation, and evaluation of VLM performance (Fig. 1). We conducted a citywide study to assess the performance of several VLMs in detecting specific residential housing conditions, including roof condition, window and door condition (e.g., broken or boarded), and facade damage, benchmarking their outputs against professional, in-person blight assessments.

\begin{figure}[p]
\centering
\includegraphics[width=\textwidth]{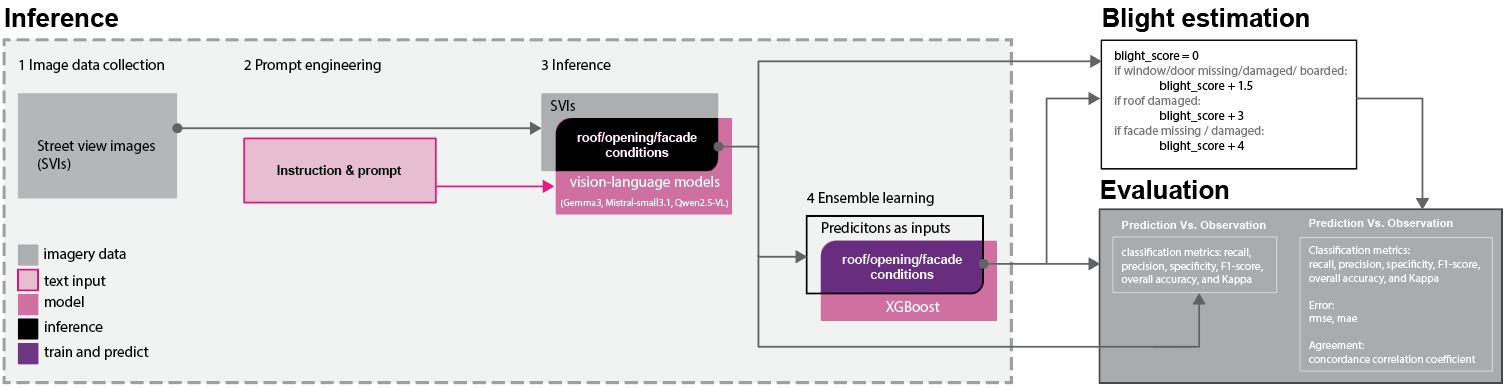}
\caption{Framework overview of the method}
\label{fig:1}
\end{figure}

\subsection{Study area and dataset}
The City of Detroit has recently begun to recover following decades of industrial decline and population loss dating to the 1970s. Yet the legacy of disinvestment remains evident, as low household incomes and housing-maintenance constraints in many neighborhoods continue to produce conditions of urban blight. According to 2018 Census estimates, Detroit had roughly 362,863 housing units and a 27\% vacancy rate (US Bureau of the Census, 2018). The Detroit Blight Removal Task Force's 2013 survey documented 35,014 residential structures requiring intervention (Erb-Downward \& Merchant, 2020). Assessing the housing stock is essential for monitoring current residential conditions, identifying areas of acute need, and informing long-term revitalization efforts (Ruggiero et al., 2020).

Our study is based on the residential structure survey of 18,886 housing units conducted by the Detroit Land Bank Authority (DLBA) in 2024 (Fig. 2). The assessor evaluated 1) roof status -  whether there were holes or substantial damage to exterior, 2) door and window openings - whether they are broken or boarded up, and 3) facade damage - whether the wall structure and surface are damaged for each residential property. Within the survey, there are corresponding street views captured from three different perspectives of each residential property, though the assessment was done in-person. Assessors made qualitative assessments of each focus area of the residence (e.g., "good condition," "small area of damage," "large area of damage"). For purposes of this study, we collapsed the original ordinal ratings into binary indicators (e.g., TRUE = "small area of damage," or "large area of damage"; FALSE = "good condition").

\begin{figure}[p]
\centering
\includegraphics[width=\textwidth]{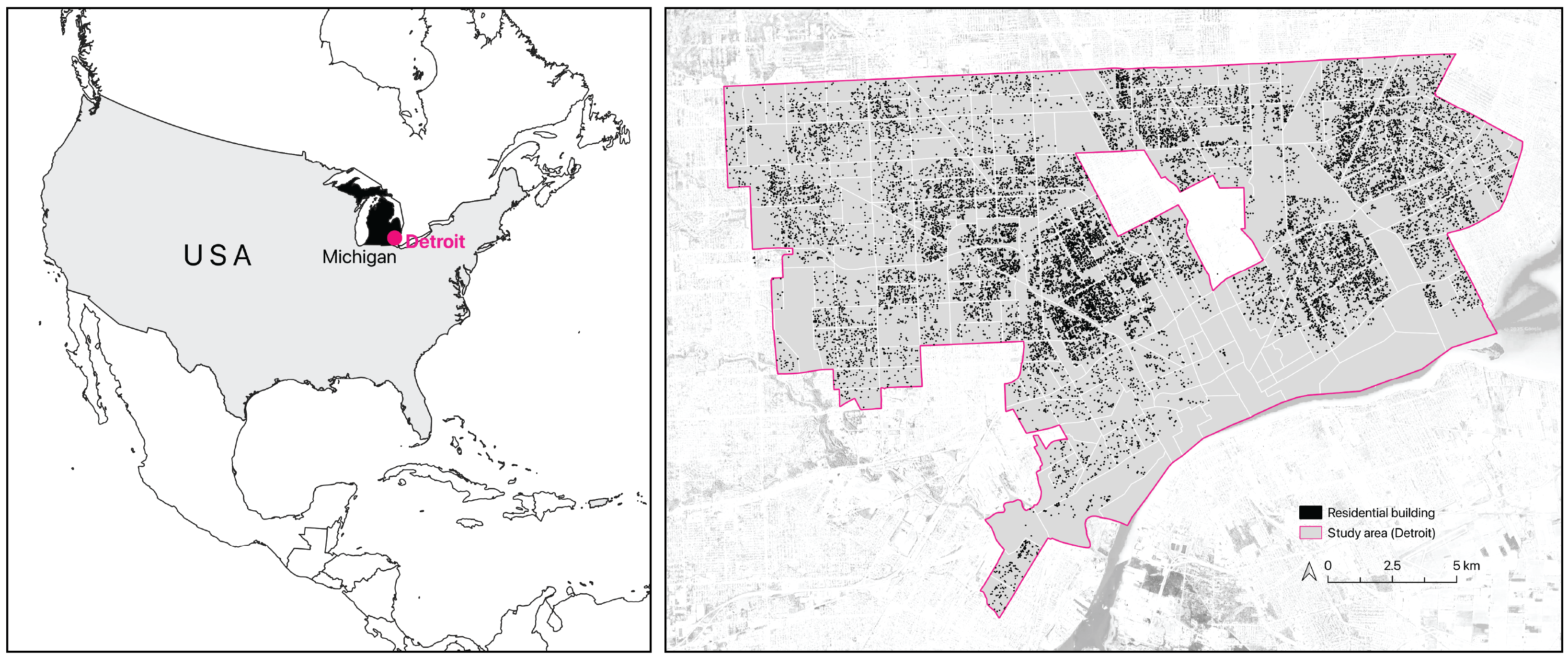}
\caption{Case study areas: the tract-level site was selected based on demographic and land value data (left); the city-level site covers most single and multi-family houses in Detroit (right)}
\label{fig:2}
\end{figure}

\subsection{Image inference and blight estimate}
\subsubsection{image inference}
To implement the image inference tasks, we used three open-source VLMs: Gemma 3, Mistral-small 3, and Qwen2.5-VL (Table 1). All models were quantized to 4 bits, reducing the computational and memory costs of running image inference and enabling their deployment for on-device use cases with a balance between precision and speed. The context length was set to 4096, which was deemed sufficient for our prompt and image inputs. To ensure the stability and consistency of structured output, we set the temperature to 0, reducing the hallucination of models during inference.

To ensure consistent and interpretable outputs from models, a structured prompt design was implemented (Appendix A). The system message provided explicit task instructions, directing the model to assess visible structural damage in residential properties. The models were instructed to return standardized responses, including (i) a binary judgment (true/false) and (ii) a probability estimate of damage occurrence on a continuous scale from 0 to 1, thereby combining categorical decision-making with calibrated uncertainty. To lead the reasoning process, the prompts instructed models to explain their observation and inference before delivering an answer, ensuring that decisions could be traced to specific visual evidence (Wei et al., 2022; Kojima et al., 2022). For each residential property, the model was applied separately to the three corresponding SVIs to generate individual inferences.

\begin{table}[p]
\caption{VLMs information}
\centering
\small
\begin{tabularx}{\textwidth}{p{0.23\textwidth} c c c c X}
\toprule
Model & Scale & Precision & Size & Context & Image size \\
\midrule
Gemma 3 & 27B & Q4\_k\_m & 17GB & 4096 & 896 \\
Mistral-small 3.1 & 24B & Q4\_k\_m & 15GB & 4096 & 1540 \\
Qwen2.5-VL & 32B & Q4\_k\_m & 21GB & 4096 & Dynamic resolution \\
\bottomrule
\end{tabularx}
\par\vspace{0.5em}\footnotesize Note: Q4\_k\_m is a 4-bit quantization with balanced precision and size.
\label{tab:vlms}
\end{table}
\subsubsection{Scoring system of blight assessment}
To quantify residential blight, we computed a parcel-level score as a severity-weighted sum across the roof, windows and doors, and fa\c{c}ade conditions, based on city assessors feedback. Scores reflect the severity of observed defects, with 0 indicating no defects. (Appendix D).  In the weighted rubric, fa\c{c}ade damage was assigned +4 and roof damage +3, reflecting their respective roles in structural stability and weatherproofing/indoor environmental quality. A missing door or window was weighted +1.5 because it poses security risks and exposes the interior to environmental hazards.

\subsection{Combining model results with ensemble stacking}
To address the accuracy limitations of individual models, we implemented a stacked ensemble in which an XGBoost meta-learner integrates the predictions from the three VLM base models (Appendix B), using the xgboost R package (Chen et al., 2019). XGBoost is a scalable tree boosting system designed for robustness when learning from large datasets, including those with correlated features produced through feature engineering (Chen et al., 2015).

Compared to simple averaging or majority voting, stacking with a gradient-boosted tree meta-learner provides greater flexibility to learn complex, nonlinear relationships among base model outputs. The inclusion of engineered features, such as probability summaries and pairwise agreement terms, allows the meta-learner to exploit not only consensus but also systematic patterns of disagreement across models (Ting \& Witten, 1999; Sagi \& Rokach, 2018; Zhou, 2025). For each observation, we constructed a feature vector consisting of: (i) the raw predictions from the three base models, (ii) summary statistics of the predicted probabilities of damage presence (mean, minimum, maximum, and standard deviation), and (iii) pairwise "agreement" and "difference" features, both on the base model predictions and on the mean probability estimates (Appendix C). These features capture both central tendencies and divergences across the models, enabling the meta-learner to exploit patterns of consensus and disagreement.

The dataset of 18,886 residential properties was split into stratified training (70\%) and testing (30\%) sets. The meta-learner was trained as a gradient-boosted tree classifier. For each hyperparameter setting, 5-fold cross-validation with early stopping was applied to select the optimal number of boosting iterations, using log loss as the evaluation criterion to prevent overfitting. Following training, predicted probabilities were converted into binary classifications by sweeping thresholds from 0.10 to 0.90 (in increments of 0.01). The threshold maximizing Cohen's Kappa (McHugh, 2012) was selected, as kappa provides a chance-corrected measure of agreement that is appropriate for imbalanced classification problems. The resulting model and tuned threshold were then fixed for all downstream ensemble predictions. Similarly, the threshold maximizing F1-score (Yang \& Liu, 1999) was selected for the resulting model that may identify more positive cases, as F1-score balances these precision and recall to assess the model's performance in identifying positive cases.

This design is particularly advantageous in the context of imbalanced classification, where naive ensembling may overweight majority-class predictions. By tuning hyperparameters through cross-validation and optimizing the decision threshold with Cohen's Kappa, the approach balances predictive accuracy with robustness to class imbalance, ensuring that the ensemble captures complementary information beyond what any single model can provide.

\subsection{Performance evaluation}
To evaluate the performance of each model, we constructed a binary confusion matrix for each residential property and utilized key performance measures for binary classifiers, including recall, precision, specificity, F1-score, overall accuracy, and Cohen's Kappa. In the binary confusion matrix, true positive (TP) denotes cases where the model correctly predicts positive instances (i.g. damaged house features); true negative (TN) denotes cases where the model correctly predicts negative instances (i.g. intact house features); false positive (FP) denotes cases where negative instances are incorrectly predicted as positive; false negative (FN) denotes cases where positive instances are incorrectly predicted as negative.

The metrics are derived from the confusion matrix and are given below:

\begin{align}
\mathrm{Recall} &= \frac{TP}{TP + FN}, &
\mathrm{Precision} &= \frac{TP}{TP + FP}, \\
\mathrm{Specificity} &= \frac{TN}{TN + FP}, &
F_1 &= \frac{2\,\mathrm{Precision}\,\mathrm{Recall}}{\mathrm{Precision}+\mathrm{Recall}}, \\
\mathrm{Accuracy} &= \frac{TP + TN}{TP + TN + FP + FN}, &
\kappa &= \frac{p_o - p_e}{1 - p_e}.
\end{align}

\begin{align}
\mathrm{MAE} &= \frac{1}{n}\sum_{i=1}^{n}\lvert x_i-y_i\rvert, &
\mathrm{RMSE} &= \sqrt{\frac{1}{n}\sum_{i=1}^{n}(x_i-y_i)^2}.
\end{align}
\section{Results}
\subsection{Performance of vision-language models}
When evaluated against professional blight assessments using the same SVI (Fig. 3a), the damaged windows and doors category (i.e., boarded-up/broken) was detected most reliably. In this task, Gemma3 achieved 0.78 accuracy, 0.86 F1, and 0.91 recall. Detection of roof damage was moderate (accuracy = 0.65--0.68), but showed a weaker precision--recall balance. Facade damage detection was the most challenging. While Qwen2.5-VL and Mistral-Small3.1 achieved high precision (0.89 - 0.93) and specificity (0.86 - 0.95), recall was relatively low (0.24 - 0.44), yielding poor overall agreement (kappa < 0.21). By model, Gemma3 produced high recall and F1-score, Qwen2.5-VL excelled in precision and specificity, and Mistral-small3.1 maintained a more balanced trade-off. Across all models, using multiple street views generally increased kappa, accuracy, recall, and F1-score, while reducing precision and specificity (Fig. 3b).

\begin{figure}[p]
\centering
\includegraphics[width=\textwidth]{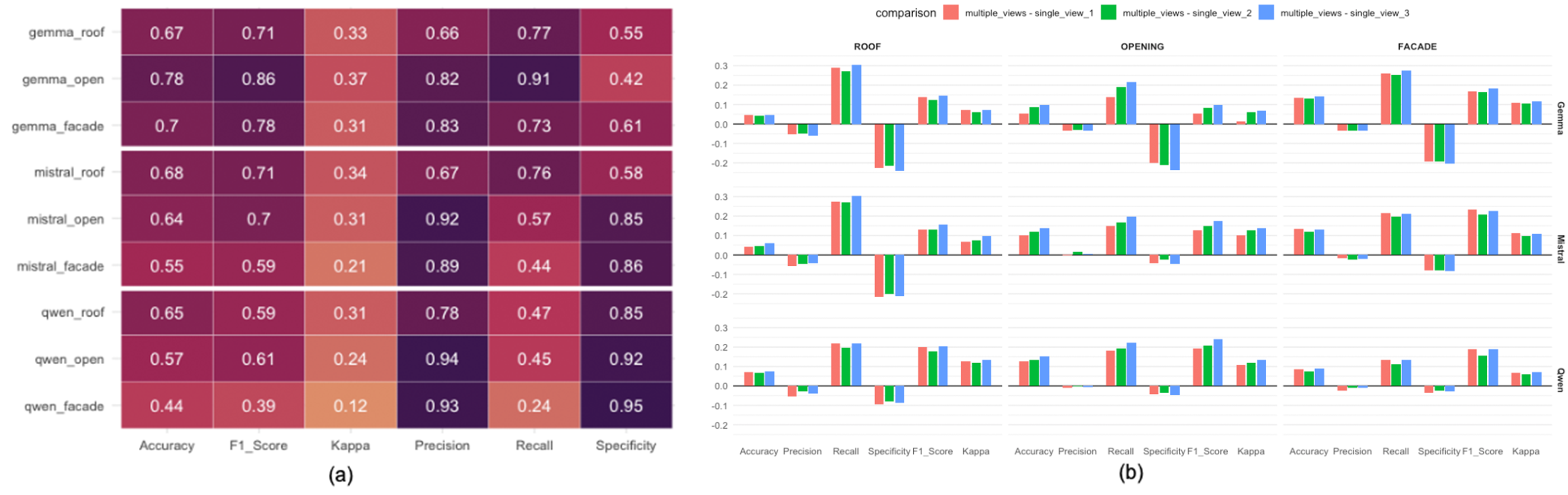}
\caption{Model performance with and without multiple views. (a) Model performance with multiple street views, and (b) differences between multiple views and every single view.}
\label{fig:3}
\end{figure}

\subsection{Assessed residential conditions and blight of individual houses}
Inter-model agreement on assessed building conditions varied by location and by category, indicating spatial heterogeneity in performance (Fig. 4, top). Although there was agreement for approximately half of the buildings, systematic differences were evident, particularly in door and window and facade evaluations, where Qwen2.5-VL and Mistral-Small-3.1 tended to align more closely than Gemma3.

Ternary maps of the average damage probability (Fig. 4, bottom) indicate that Gemma3 and Mistral-small3.1 generally returned similar probability ranges across all residential conditions. Mistral-Small-3.1 tended to assign higher predicted probabilities of damage, whereas Qwen2.5-VL produced a wider spread skewed toward lower probabilities (more blue points).  Roof-damage predictions showed reduced between-model variance, with probabilities concentrated in a narrow band (dominant purple/gray points).

\begin{figure}[p]
\centering
\includegraphics[width=\textwidth]{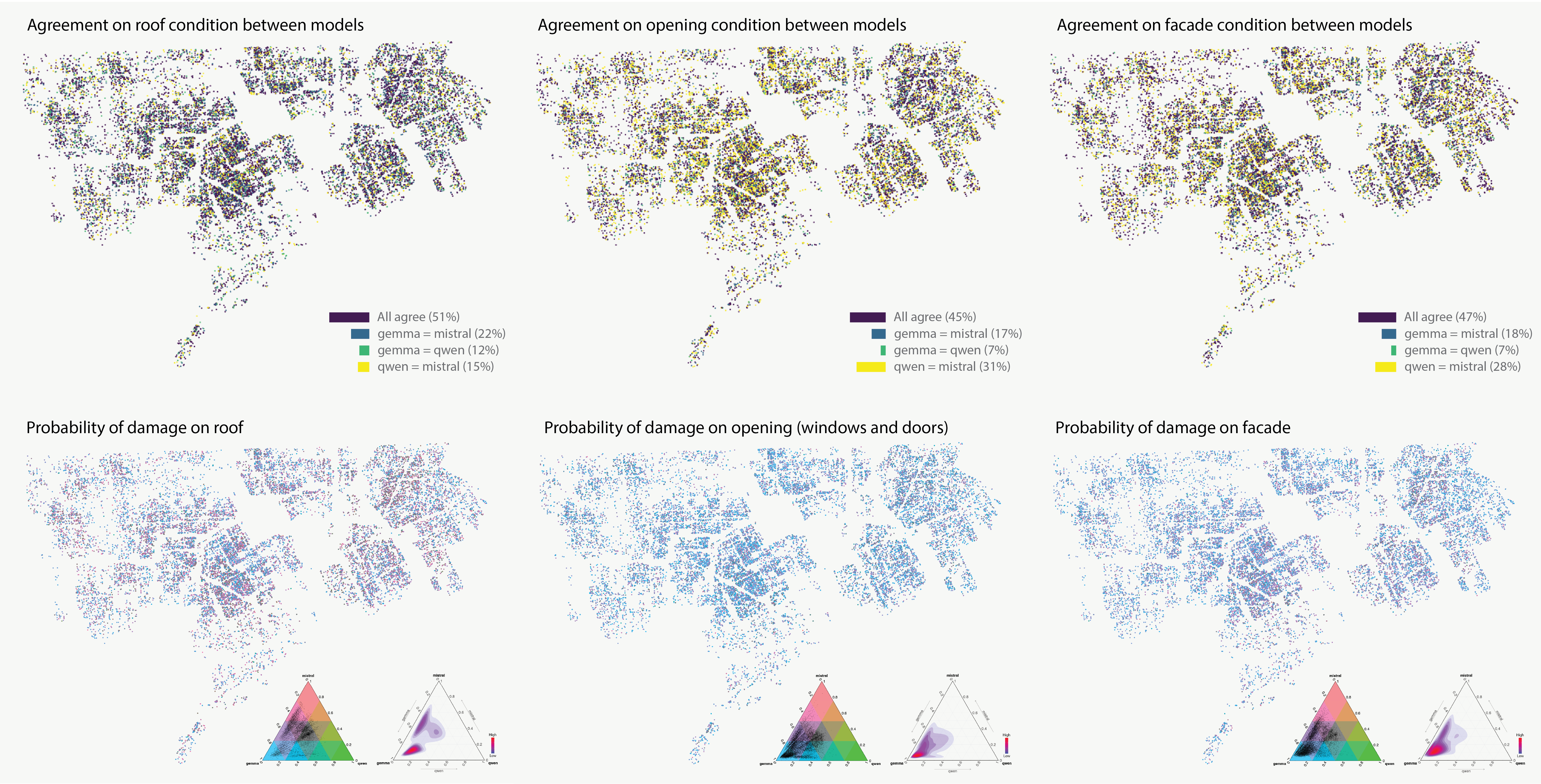}
\caption{Agreement on residential conditions (top) and average damage probability (bottom) between models (Note: ternary diagrams for average damage probability are provided in Appendix E). For roof condition, all three models agreed on 51\% of buildings, while partially there were partial agreements among Gemma--Mistral (22\%), Qwen--Mistral (15\%), and Gemma--Qwen (12\%). Agreement was lowest for the detection of damage in windows and doors (opening), with 45\% full consensus and a dominant pairwise alignment between Qwen2.5-VL and Mistral-small3.1 (31\%). Facade condition showed 47\% full agreement, again with Qwen--Mistral as the strongest pairwise combination (28\%) and Gemma--Qwen as the weakest (7\%).}
\label{fig:4}
\end{figure}

Estimates of residential blight varied across models (Fig. 5a). Gemma3 predicted the highest baseline levels, with widespread medium-to-high blight values (green-to-yellow) distributed across much of the study area. In contrast, Mistral-small3.1 produced more spatially concentrated patterns, identifying distinct hot spots of severe blight, particularly in central neighborhoods. Qwen2.5-VL generated the most conservative estimates, with a majority of properties falling into lower blight and fewer classified at the very high end. Despite differences in intensity, all three models consistently highlighted similar clusters of distressed housing (Fig. 4c). This indicates shared detection of broad spatial patterns, with model-specific sensitivities to building conditions that mirror their per-condition performance.
Summarizing model performance, Gemma3 shows the most symmetric error pattern and the lowest overall error (RMSE = 3.37, MAE = 2.30) with residuals tightly centered near zero, indicating relatively unbiased estimates (Fig. 5b). Mistral-small3.1 shows a wider error spread (RMSE = 3.76, MAE = 2.81) and a slight tendency toward underestimation. Qwen2.5-VL performed the weakest (RMSE = 4.80, MAE = 3.77), with a broad and left-skewed error distribution, suggesting a consistent bias toward lower blight scores.

\begin{figure}[p]
\centering
\includegraphics[width=\textwidth]{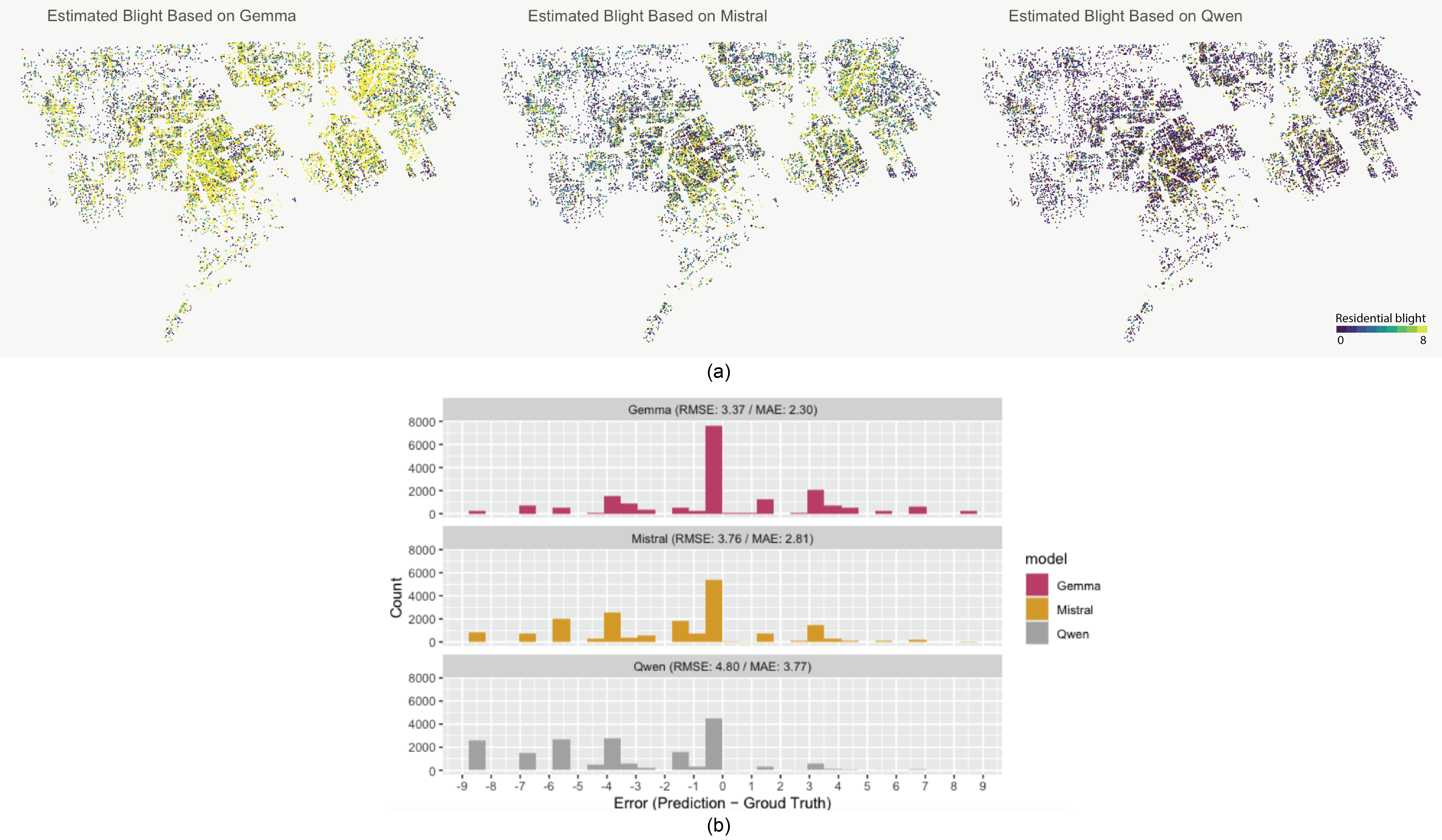}
\caption{Distribution of (a) estimated blight and (b) estimation error}
\label{fig:5}
\end{figure}

\subsection{Comparison of ensemble learner and base models}
The comparison of the ensemble learner's performance against the three base models across roof, window and door, and facade conditions reveals that the ensemble achieves balanced improvements across most tasks. With thresholds chosen to maximize kappa (Fig. 6a,b), the ensemble model attains higher accuracy (0.7 - 0.77), F1-score (0.71 - 0.84), and kappa (0.41 - 0.43) across all residential conditions compared to the base models, with a slight drop in accuracy and F1-score compared to Gemma3. In addition, deltas also show consistent kappa gains--most clearly for windows and doors and facade conditions with small drops in precision and specificity. When thresholds maximize F1 (Fig. 6c,d), the ensemble model drives recall much higher (0.83 - 0.95) at the cost of lower precision (0.68 - 0.78) and specificity (0.27 - 0.55) compared to the kappa-oriented ensemble model. The deltas also highlight large recall improvements, especially for wall condition (0.3 - 0.5), versus the base models, while the improvement in kappa and accuracy is not significant. Overall, the ensemble models provided the best prediction on windows and doors and facade conditions, with the kappa-vs-F1 threshold objective controlling the precision--recall--specificity trade-off.

\begin{figure}[p]
\centering
\includegraphics[width=\textwidth]{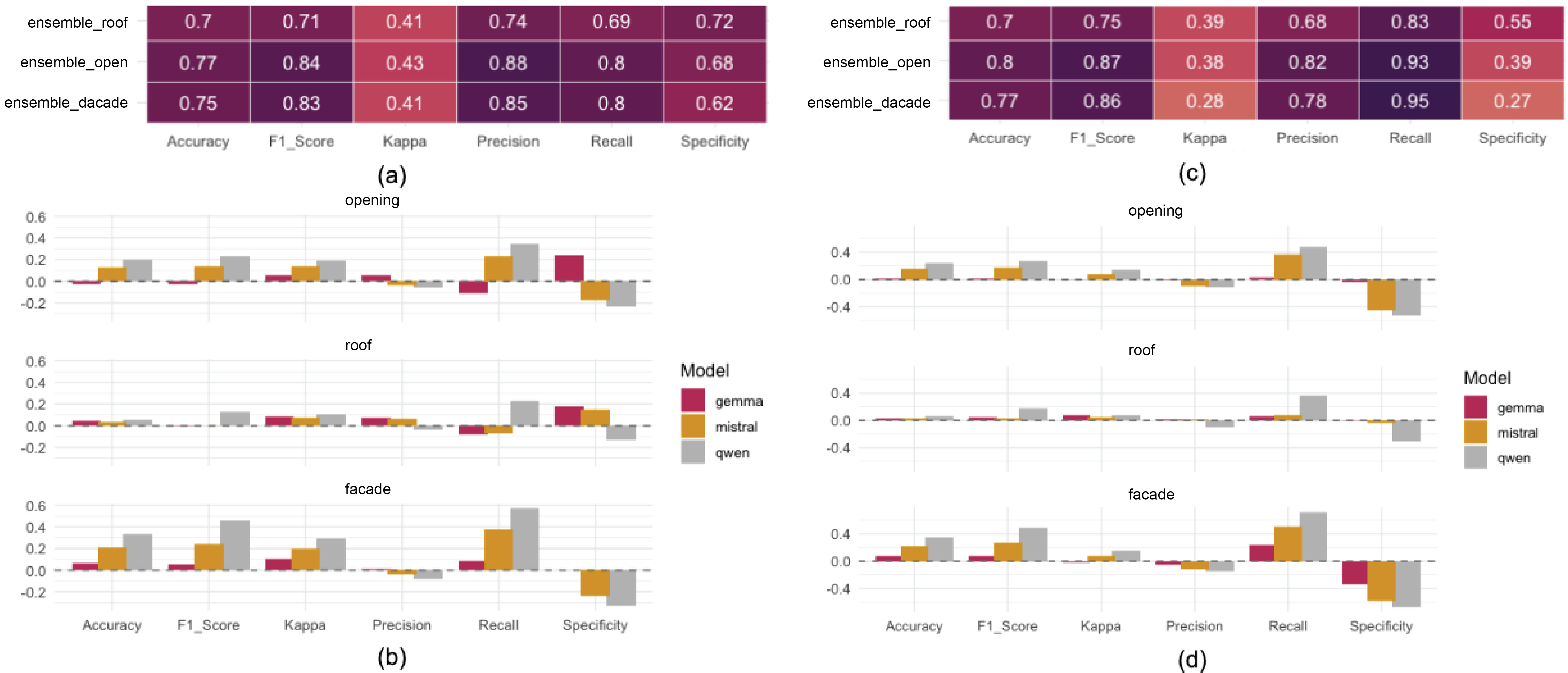}
\caption{Comparison between ensemble and base models based on the test set: the performance of the ensemble learner and base models with threshold selected for maximizing kappa (a) and for maximizing F1-score (c); (b) the delta between the performance of the ensemble learner and base models with threshold selected for maximizing kappa (b) and for maximizing F1-score (d)}
\label{fig:6}
\end{figure}

The ensemble model achieves the lowest RMSE (3.01, 3.05) and MAE (1.97), with errors distributed more tightly around zero compared to the base models and the highest CCC (0.537, 95\% CI: 0.519--0.555), reflecting both strong correlation and reduced bias relative to the best-fit line. In addition, the F1-score-oriented threshold tends to produce more overestimated cases, and fewer underestimated cases compared to the ensemble model for maximizing kappa, with a lower CCC (0.469, 95\% CI: 0.450--0.488), while the regression slope is above the best-fit line. This indicates better performance in identifying houses with potential blight and more reliable scaling across the full range of blight severity (Fig. 7).

\begin{figure}[p]
\centering
\includegraphics[width=\textwidth]{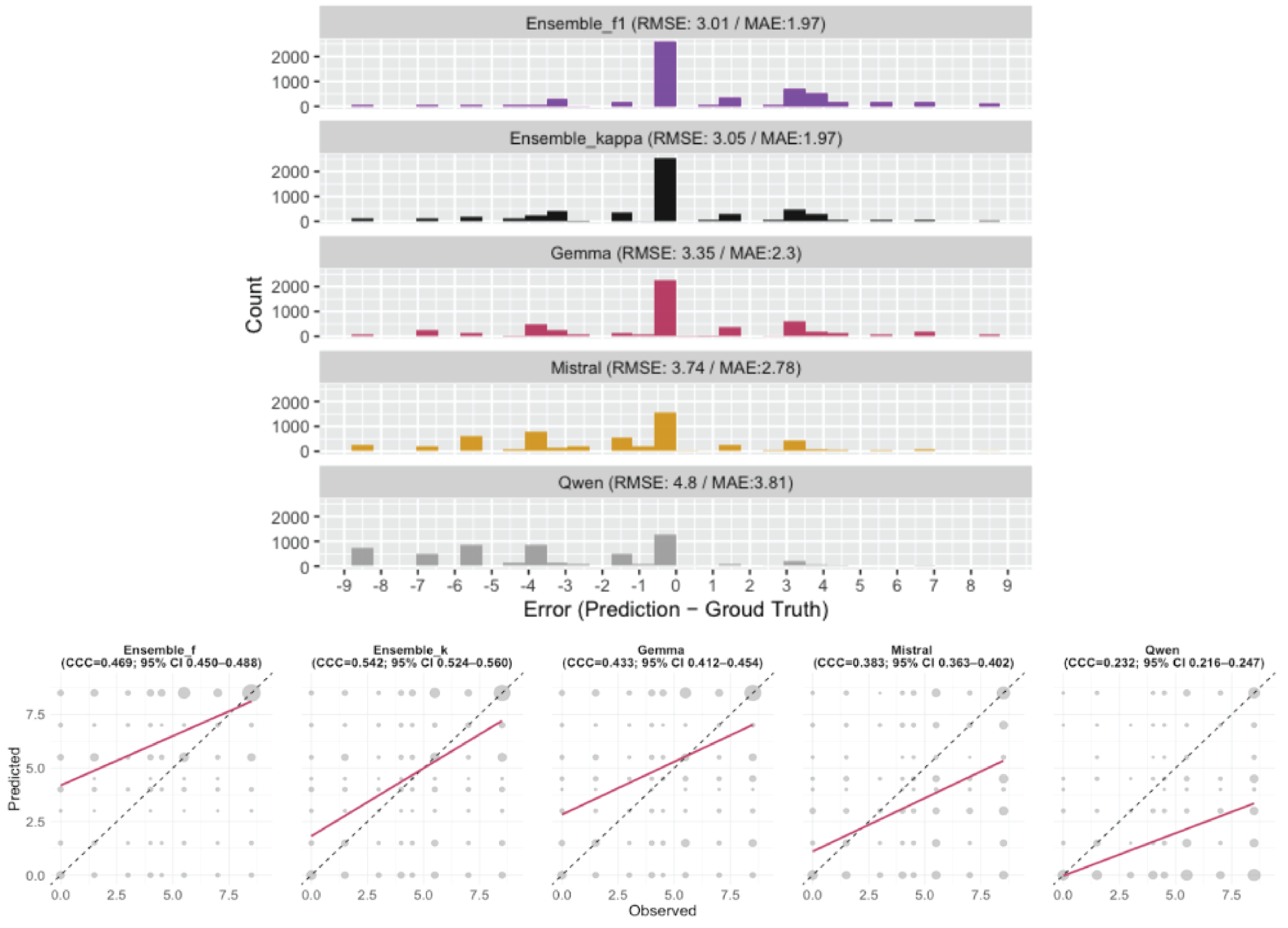}
\caption{Error distribution (top) and concordance correlation coefficient (bottom): In concordance correlation coefficient results, the larger circle sizes around mid-range values suggest that errors accumulate where most properties are concentrated, and the regression slope is closer to the best-fit line (45-degree dashed line) when there is a stronger agreement between two sets of data points}
\label{fig:7}
\end{figure}

\section{Discussion}
\subsection{Characteristics of models with multiple perspectives}
In this paper, we test a scalable framework for mapping residential blight in Detroit by leveraging SVIs and open-source large VLMs. Based on the citywide investigation, we found that the three VLMs demonstrate distinct patterns in identifying different types of residential damage. Gemma3 performed the best at identifying broken or boarded windows and doors; however, it tends to overpredict damage, yielding more false positives. Qwen2.5-VL behaved conservatively at these vision classification tasks, exhibiting high precision but low recall. In practice, it rarely flags intact structures yet frequently overlooks subtle or partially occluded defects, especially on facade conditions. By contrast, Mistral-Small-3.1 delivers a more balanced profile across categories, with relatively stable precision and recall, though its overall agreement with human annotations is modestly lower, suggesting calibration and threshold tuning could further align predictions with assessor judgments.

We also found that using multiple street-view perspectives and ensembling across models improved the robustness of damage detection for roofs, door and window openings, and fa\c{c}ades in the city-scale case study. Although the three base VLMs have distinct inference behaviors, they achieved substantial agreement on a large number of housing units across all residential conditions. Meanwhile, we found that the estimated probabilities of housing damage by different models demonstrated similar spatial patterns, showing the high-risk zones or hot spots in Detroit. Urban planners and managers can use these maps to zoom into neighborhoods with relatively more agreement to further investigate individual properties.
This suggests that our approach can complement manual assessments by highlighting hotspots and guiding assessors to target field inspections and allocate resources more efficiently for an otherwise costly, labor-intensive task.

\subsection{Contribution of ensemble meta-learner}
Our results demonstrate that stacking via XGBoost can produce more reliable predictions than relying on any single model, outperforming each base model in both error reduction and agreement strength. From the ensemble learners trained by XGBoost, we further explored the importance of input signals. Across all three categories, the mean predicted probabilities from the base models dominate feature importance (Fig. 8), indicating that the ensemble learners primarily rely on the collective damage probability of the base models. The minimum and maximum probabilities also have a moderate influence, reflecting that the ensemble learners capture both consensus and extremity in base predictions. In contrast, agreement features and vote-based indicators contribute less.

\begin{figure}[p]
\centering
\includegraphics[width=\textwidth]{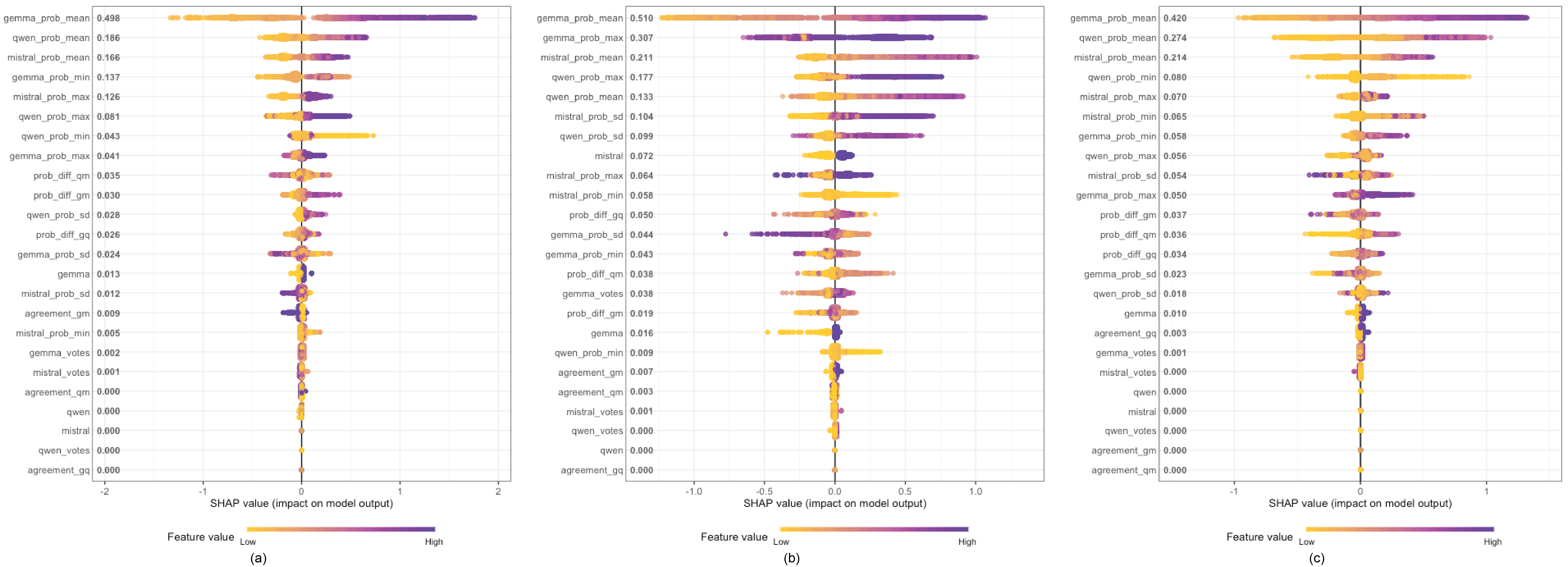}
\caption{SHAP value of ensemble models (a: roof; b: opening; c: facade)}
\label{fig:8}
\end{figure}

\subsection{Application to assessing urban blight in residential areas of Detroit}
Our results have meaningful real-world applications. Housing-condition maps can identify parcels with visible deterioration (binary indicators), while probability surfaces quantify uncertainty and the relative severity of blight. Together, these layers can inform urban planning and neighborhood revitalization by distinguishing clearly deteriorated structures from those in earlier stages of decline.

Moreover, our approach is broadly applicable. Classification thresholds can be tuned to the objective, for example, higher recall for broad screening or higher precision for enforcement. While comprehensive, in-person inspections are often necessary to assess and manage housing conditions, interim estimates can guide triage and planning. Using these estimates helps prevent systematic underestimation of blight and reduces the likelihood that high-need properties are missed. Accordingly, an F1-score-oriented threshold needs to be employed to gain a more aggressive blight estimation. Therefore, the estimation of residential blight can serve as a low-cost and quick update complementary document for the Detroit Land Bank Authority to review and compare to the blight survey.

Beyond blight detection, the methodological framework is adaptable to related urban environments characterized by visible cues. Potential applications include evaluating outdoor maintenance (e.g., Peng et al., 2025), housing (e.g., Yue et al., 2022), and storefront vacancy (e.g., Li \& Long, 2024). Each of these domains involves spatially distributed visual cues that can be captured from street-level imagery and analyzed through VLMs and similar ensemble-based inference.

\subsection{Limitations and future research}
While the ensemble approach can improve the capacity of open-source VLMs, the model performance indicates moderate agreement between prepositions and observations, indicating the need for continued methodological refinement to achieve higher accuracy and reliability. This study points to several promising research directions. First, expanding the number of VLMs for the ensemble stacking could increase the diversity of inference patterns and improve decision robustness. Beyond the three models applied in this study, other open-source VLMs such as Intern-VL and Llama could provide valuable complementary perspectives for evaluating residential structures. Second, the influence of inference parameters--particularly the temperature setting--should be systematically examined, as it plays an important role in controlling response variability but remains poorly understood in the context of residential condition assessment. Third, open-source VLMs need to be fine-tuned on balanced datasets for more accurate estimation of residential damage. Even if the ensemble models can gain higher accuracy, the agreement between the prediction and observation is still not substantial, as kappa is lower than 0.6 for the residential conditions. Finally, alternative strategies such as one-shot or few-shot learning may offer efficient pathways to enhance performance without extensive retraining, especially for unfinetuned open-source models.

\section{Conclusion}
This study demonstrates the potential of using open-source VLMs for assessing residential building conditions at scale using multiple views and the advantages of integrating multiple model predictions through ensemble learning. Moreover, it proved accurate compared to trained in-person assessors. By combining structured prompting, probability-based feature engineering, and an XGBoost meta-learner, the proposed framework generates more accurate, interpretable, and spatially coherent assessments of residential blight than any single model alone. The results show that (i) multiple SVIs can enhance the robustness of detecting roof, opening, and wall damages; (ii) the three open-source VLMs exhibit distinct inference strengths across damage types--Gemma3 performs best for openings but tends to overpredict, Qwen2.5-VL is conservative with high precision but low recall, and Mistral-small3.1 achieves balanced yet moderate performance; and (iii) the ensemble learner outperforms all individual models, enhancing robustness across residential condition categories.

Applied to Detroit, the binary and probability maps provide actionable information for city planners and managers, enabling targeted reinvestment, prioritization of inspection, and evidence-based policy interventions. The probabilistic outputs of the ensemble learner allow for flexible threshold adjustments to balance recall and precision depending on operational goals. As a result, the framework can serve as a low-cost, regularly updatable complement to traditional housing surveys, offering an efficient means to support the Detroit Land Bank Authority and similar institutions in tracking and managing housing stock conditions.

More broadly, this work contributes to the growing field of AI-assisted urban analytics by demonstrating that VLMs--when integrated through ensemble learning--can bridge the gap between human interpretive capacity and machine scalability. Moreover, the moderate agreement between predictions and ground truth suggests that further refinement is necessary. Future research should explore (i) expanding the number and diversity of VLMs to increase ensemble stability; (ii) investigating the role of inference temperature in shaping model performance; (iii) fine-tuning open-source VLMs on balanced, domain-specific datasets or applying few-shot learning strategies to improve generalization.

\appendix
\clearpage
\section{System prompts and question prompts for two case studies}
\begin{figure}[H]
\centering
\includegraphics[width=\textwidth,height=0.78\textheight,keepaspectratio]{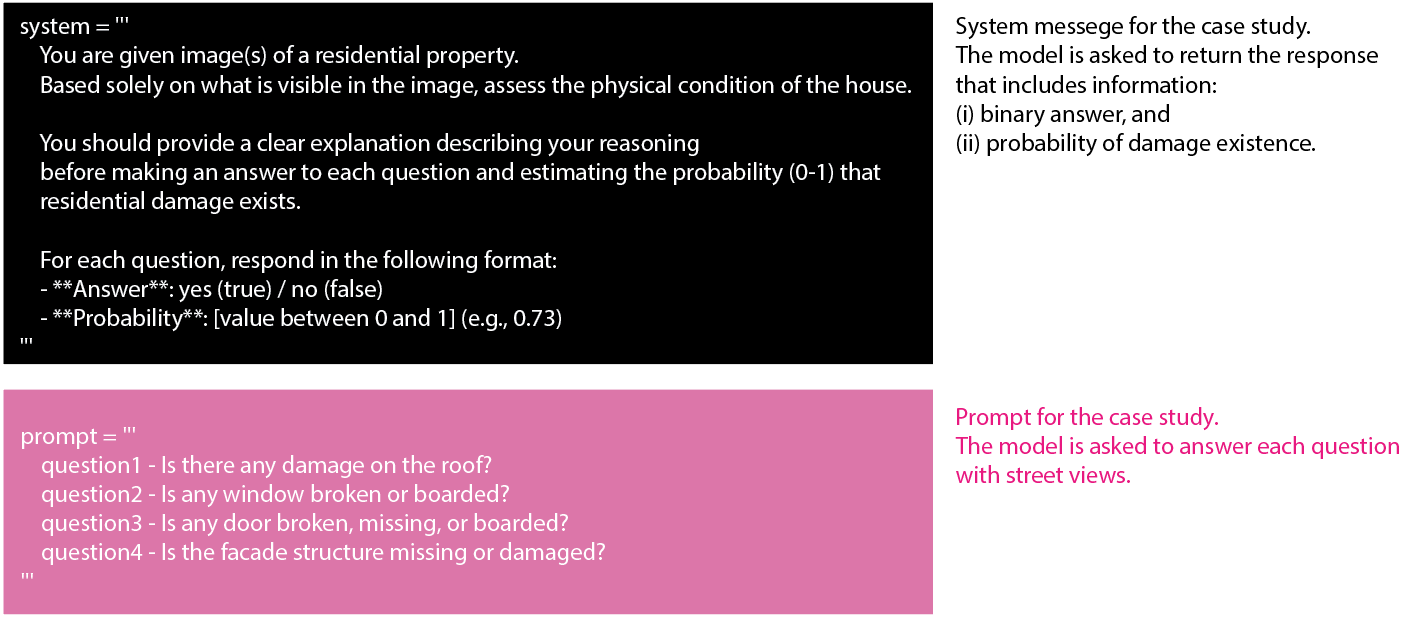}
\label{app:a}
\end{figure}

\clearpage
\section{Ensemble approach of integrating the predictions of base models for each condition}
\begin{figure}[H]
\centering
\includegraphics[width=\textwidth,height=0.78\textheight,keepaspectratio]{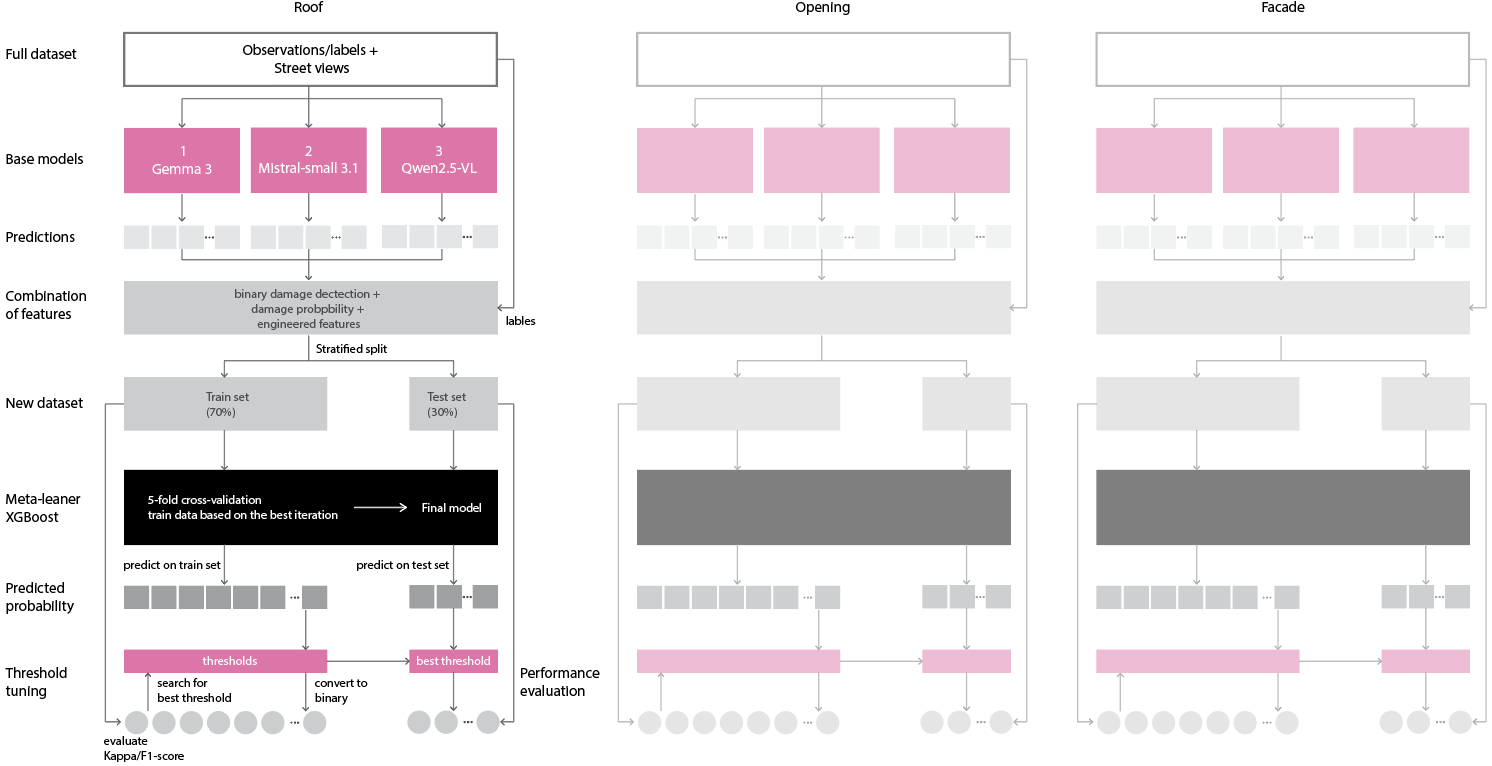}
\label{app:b}
\end{figure}

\clearpage
\section{Input features for stacking learning}
{\small
\begin{longtable}{p{0.28\textwidth} p{0.30\textwidth} p{0.34\textwidth}}
\label{tab:stacking-features}\\
\toprule
Features & Formula & Description \\
\midrule
\endfirsthead
\toprule
Features & Formula & Description \\
\midrule
\endhead
\begin{tabular}[t]{@{}l@{}}Binary prediction of Gemma ($y_g$)\\Binary prediction of Qwen ($y_q$)\\Binary prediction of Mistral ($y_m$)\end{tabular} & $y_j=\begin{cases}0, & 1\notin\{x_{1j},x_{2j},x_{3j}\}\\1, & 1\in\{x_{1j},x_{2j},x_{3j}\}\end{cases},\ j\in\{g,q,m\}$ & For each property and damage category, model $j$ produces binary predictions $x$. If any of the three street views is predicted as positive, the property-level outcome is set to 1; otherwise, it is 0. \\
\addlinespace
\begin{tabular}[t]{@{}l@{}}Total votes of Gemma ($v_g$)\\Total votes of Qwen ($v_q$)\\Total votes of Mistral ($v_m$)\end{tabular} & $v_j=\sum_{i=1}^{3}x_{ij}$ & This sums the binary predictions across all street views for a property. For example, if Gemma predicts damage in 2 out of 3 street views, then $v_g=2$. \\
\addlinespace
\begin{tabular}[t]{@{}l@{}}Gemma probability mean ($\mu_g$)\\Qwen probability mean ($\mu_q$)\\Mistral probability mean ($\mu_m$)\end{tabular} & $\mu_j=\frac{1}{3}\sum_{i=1}^{3}p_{ij}$ & Each model outputs a probability $p$ of damage for each street view. The formula averages these three probabilities to obtain the mean predicted probability for the property. \\
\addlinespace
\begin{tabular}[t]{@{}l@{}}Gemma min probability ($\min_g$)\\Qwen min probability ($\min_q$)\\Mistral min probability ($\min_m$)\end{tabular} & $\min_j=\min\{p_{1j},p_{2j},p_{3j}\}$ & This takes the lowest probability from the three street views, capturing the least confident estimate of damage. \\
\addlinespace
\begin{tabular}[t]{@{}l@{}}Gemma max probability ($\max_g$)\\Qwen max probability ($\max_q$)\\Mistral max probability ($\max_m$)\end{tabular} & $\max_j=\max\{p_{1j},p_{2j},p_{3j}\}$ & This takes the highest probability of damage among the three street views, reflecting the strongest evidence of damage predicted by a model. \\
\addlinespace
\begin{tabular}[t]{@{}l@{}}Gemma sd probability ($\sigma_g$)\\Qwen sd probability ($\sigma_q$)\\Mistral sd probability ($\sigma_m$)\end{tabular} & $\sigma_j=\operatorname{SD}\{p_{1j},p_{2j},p_{3j}\}$ & This measures how much the predicted probabilities vary across the three street views. A higher standard deviation indicates less consistency across angles. \\
\addlinespace
\begin{tabular}[t]{@{}l@{}}Agreement between Gemma and Qwen ($\alpha_{gq}$)\\Agreement between Gemma and Mistral ($\alpha_{gm}$)\\Agreement between Qwen and Mistral ($\alpha_{qm}$)\end{tabular} & $\alpha_{jk}=|y_j-y_k|,\ j,k\in\{g,q,m\}$ & This calculates the absolute difference between binary predictions for two models. If both models agree, the value is 0; if they disagree, the value is 1. \\
\addlinespace
\begin{tabular}[t]{@{}l@{}}Difference between Gemma and Qwen ($\beta_{gq}$)\\Difference between Gemma and Mistral ($\beta_{gm}$)\\Difference between Qwen and Mistral ($\beta_{qm}$)\end{tabular} & $\beta_{jk}=|\mu_j-\mu_k|$ & This takes the absolute difference between the average probabilities of two models, showing how far apart their overall probability estimates are for the same property. \\
\bottomrule
\end{longtable}
}
\section{Scoring system for blight estimation}
\begin{table}[H]
\centering
\begin{tabular}{lc}
\toprule
Condition & Score \\
\midrule
Roof damaged & +3 \\
Opening damaged (window or door damage) & +1.5 \\
Facade damaged & +4 \\
\bottomrule
\end{tabular}
\label{tab:blight-score}
\end{table}
\clearpage
\section{Ternary maps of damage probability by VLMs for residential conditions}
\begin{figure}[H]
\centering
\includegraphics[width=\textwidth,height=0.78\textheight,keepaspectratio]{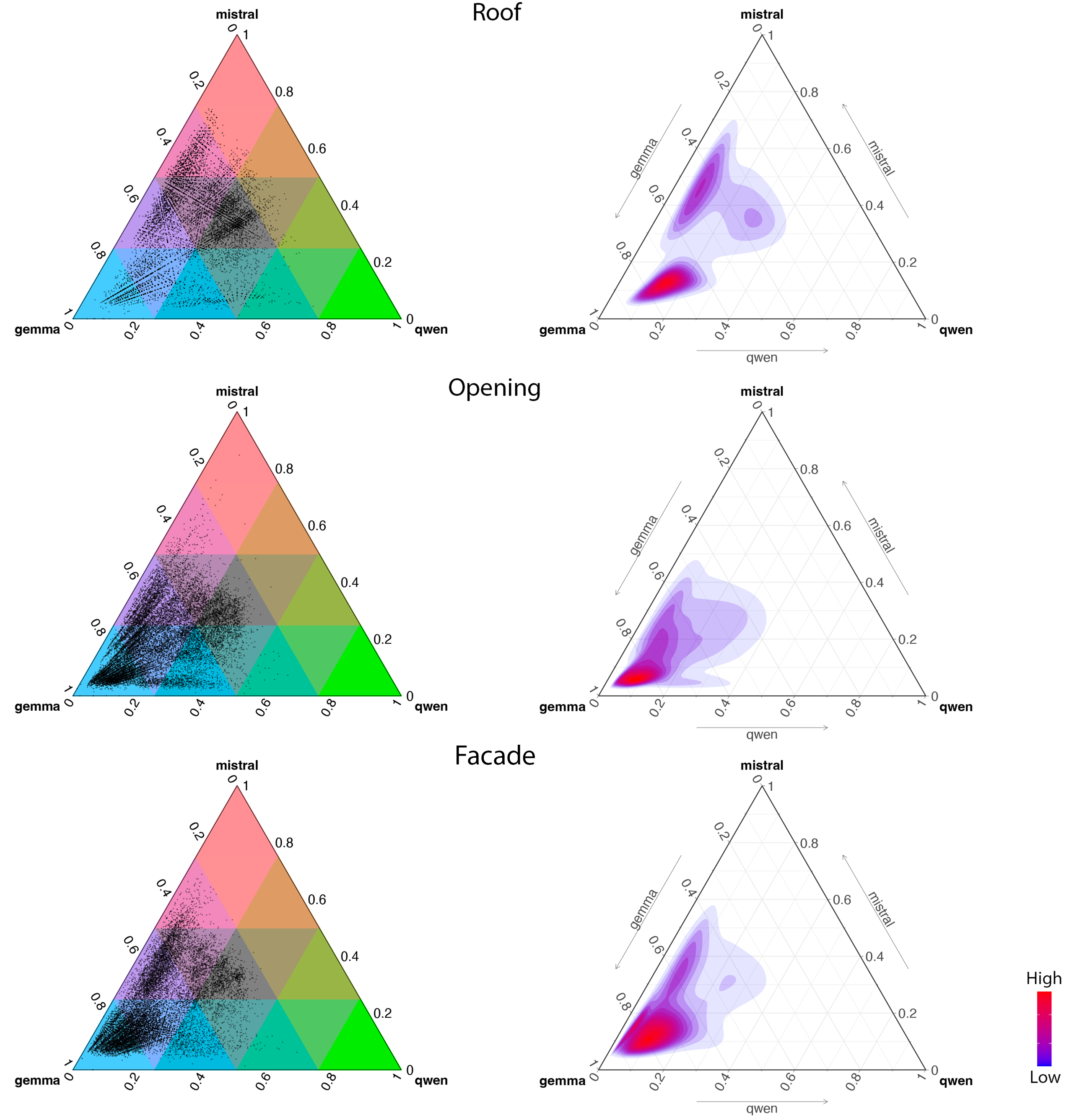}
\label{app:e}
\end{figure}

\section*{References}
\begin{enumerate}
\item Alam, F., Ofli, F., \& Imran, M. (2018, June). Crisismmd: Multimodal twitter datasets from natural disasters. In Proceedings of the international AAAI conference on web and social media (Vol. 12, No. 1).

\item Atkinson, R. (2004). The evidence on the impact of gentrification: new lessons for the urban renaissance?. European journal of housing policy, 4(1), 107-131.

\item Bai, S., Chen, K., Liu, X., Wang, J., Ge, W., Song, S., ... \& Lin, J. (2025). Qwen2. 5-vl technical report. arXiv preprint arXiv:2502.13923.

\item Beers, A., Daley, C., McLaughlin, I., \& Pavlek, G. (2011). Quick guide: New tools to address blight and abandonment. Pennsylvania: The Housing Alliance of Pennsylvania.

\item Bibri, S. E. (2019). On the sustainability of smart and smarter cities in the era of big data: an interdisciplinary and transdisciplinary literature review. Journal of Big Data, 6(1), 25.

\item Bieri, V., Zamboni, M., Blumer, N. S., Chen, Q., \& Engelmann, F. (2025, February). Opencity3d: What do vision-language models know about urban environments?. In 2025 IEEE/CVF Winter Conference on Applications of Computer Vision (WACV) (pp. 5147-5155). IEEE.

\item Bonnefoy, X. (2007). Inadequate housing and health: an overview. International journal of environment and pollution, 30(3-4), 411-429.

\item Board on Health Promotion, Disease Prevention, \& Committee on Damp Indoor Spaces. (2004). Damp indoor spaces and health.

\item Buckland, M., \& Gey, F. (1994). The relationship between recall and precision. Journal of the American society for information science, 45(1), 12-19.

\item Cao, J., Wang, H., Li, J., Tian, Q., \& Niyogi, D. (2022). Improving the forecasting of winter wheat yields in Northern China with machine learning--dynamical hybrid subseasonal-to-seasonal ensemble prediction. Remote Sensing, 14(7), 1707.

\item Chai, T., \& Draxler, R. R. (2014). Root mean square error (RMSE) or mean absolute error (MAE). Geoscientific model development discussions, 7(1), 1525-1534.

\item Chen, T., He, T., Benesty, M., \& Khotilovich, V. (2019). Package 'xgboost'. R version, 90(1-66), 40.

\item Chen, T., He, T., Benesty, M., Khotilovich, V., Tang, Y., Cho, H., ... \& Zhou, T. (2015). Xgboost: extreme gradient boosting. R package version 0.4-2, 1(4), 1-4.

\item Chen, Z., Li, J., Chen, P., Li, Z., Sun, K., Luo, Y., ... \& Yu, P. S. (2025). Harnessing multiple large language models: A survey on llm ensemble. arXiv preprint arXiv:2502.18036.

\item Cho, D., Yoo, C., Im, J., Lee, Y., \& Lee, J. (2020). Improvement of spatial interpolation accuracy of daily maximum air temperature in urban areas using a stacking ensemble technique. GIScience \& Remote Sensing, 57(5), 633-649.

\item Detroit Land Bank Authority. (2025). City Council quarterly report: Q1 FY25. Detroit Land Bank Authority. https://dlba-production-bucket.s3.us-east-2.amazonaws.com/City\_Council\_Quarterly\_Report/DLBA+Q1+FY25+CCQR+FINAL.pdf

\item Dona, M. A. M., Cabrero-Daniel, B., Yu, Y., \& Berger, C. (2024). Evaluating and enhancing trustworthiness of LLMs in perception tasks. arXiv preprint arXiv:2408.01433. https://arxiv.org/abs/2408.01433

\item Erb-Downward, J., \& Merchant, S. (2020). Losing home: Housing instability \& availability in Detroit. Retrieved from Poverty Solutions website: https://poverty. umich. edu/files/2020/05/200358\_Poverty-Solutions\_Detroit-Housing-Instability-policy-brief\_051120. Pdf.

\item Feng, J., Du, Y., Liu, T., Guo, S., Lin, Y., \& Li, Y. (2024). CityGPT: Empowering urban spatial cognition of large language models. arXiv preprint arXiv:2406.13948. https://arxiv.org/abs/2406.13948

\item Ghosh, S. M., Behera, M. D., Jagadish, B., Das, A. K., \& Mishra, D. R. (2021). A novel approach for estimation of aboveground biomass of a carbon-rich mangrove site in India. Journal of Environmental Management, 292, 112816.

\item Gillespie, L. E., Ruffley, M., \& Exposito-Alonso, M. (2024). Deep learning models map rapid plant species changes from citizen science and remote sensing data. Proceedings of the National Academy of Sciences, 121(37), e2318296121.

\item Healey, S. P., Cohen, W. B., Yang, Z., Brewer, C. K., Brooks, E. B., Gorelick, N., ... \& Zhu, Z. (2018). Mapping forest change using stacked generalization: An ensemble approach. Remote Sensing of Environment, 204, 717-728.

\item Huang, Z., Qi, H., Kang, C., Su, Y., \& Liu, Y. (2020). An ensemble learning approach for urban land use mapping based on remote sensing imagery and social sensing data. Remote Sensing, 12(19), 3254.

\item Jiang, Y., Chao, Q., Chen, Y., Li, X., Liu, S., \& Cong, G. (2024). UrbanLLM: Autonomous urban activity planning and management with large language models. arXiv preprint arXiv:2406.12360. https://arxiv.org/abs/2406.12360

\item Jiang, Y., \& Sun, P. (2024). Does shrinkage have an impact on urban livability? An empirical analysis from Northeast China. Sustainable Cities and Society, 113, 105725.

\item Kojima, T., Gu, S. S., Reid, M., Matsuo, Y., \& Iwasawa, Y. (2022). Large language models are zero-shot reasoners. Advances in neural information processing systems, 35, 22199-22213.

\item Lawrence, I., \& Lin, K. (1989). A concordance correlation coefficient to evaluate reproducibility. Biometrics, 255-268.

\item Li, Y., \& Long, Y. (2024). Inferring storefront vacancy using mobile sensing images and computer vision approaches. Computers, Environment and Urban Systems, 108, 102071.

\item Li, Z., Xia, L., Tang, J., Xu, Y., Shi, L., Xia, L., Yin, D., \& Huang, C. (2024). UrbanGPT: Spatio-temporal large language models. In Proceedings of the 30th ACM SIGKDD Conference on Knowledge Discovery and Data Mining (pp. 5351--5362). Association for Computing Machinery. https://doi.org/10.1145/3637528.3671578

\item Liang, X., Brainerd, B., Hicks, T., \& Andris, C. (2024). Lessons from a human-in-the-loop machine learning approach for identifying vacant, abandoned, and deteriorated properties in Savannah, Georgia. Journal of Planning Education and Research, 0739456X241273945.

\item Liang, X., Xie, J., Zhao, T., Stouffs, R., \& Biljecki, F. (2025). OpenFACADES: An Open Framework for Architectural Caption and Attribute Data Enrichment via Street View Imagery. arXiv preprint arXiv:2504.02866.

\item Lv, L., Chen, T., Dou, J., \& Plaza, A. (2022). A hybrid ensemble-based deep-learning framework for landslide susceptibility mapping. International Journal of Applied Earth Observation and Geoinformation, 108, 102713.

\item Malekzadeh, M., Willberg, E., Torkko, J., \& Toivonen, T. (2025). Urban attractiveness according to ChatGPT: Contrasting AI and human insights. Computers, Environment and Urban Systems, 117, 102243.

\item McHugh, M. L. (2012). Interrater reliability: the kappa statistic. Biochemia medica, 22(3), 276-282.

\item Mistral AI. (2025, March 17). Mistral Small 3.1. Mistral AI. https://mistral.ai/news/mistral-small-3-1

\item Neidert, L., Farley, R., \& Morenoff, J. (2025). How Census Undercount Became a Civil Rights Issue and Why It Is Increasingly Important. RSF: The Russell Sage Foundation Journal of the Social Sciences, 11(1), 26--43.

\item Niimi, J. (2025, July). A simple ensemble strategy for llm inference: Towards more stable text classification. In International Conference on Applications of Natural Language to Information Systems (pp. 189-199). Cham: Springer Nature Switzerland.

\item Ochodo, C., Ndetei, D. M., Moturi, W. N., \& Otieno, J. O. (2014). External built residential environment characteristics that affect mental health of adults. Journal of Urban Health, 91, 908-927.

\item Peng, Q., Zhao, G., \& Ye, X. (2025). Assessing the impact of maintenance condition on multifamily rents: an integrated approach of machine learning and hedonic modelling. Journal of Housing and the Built Environment, 1-18.

\item Pevalin, D. J., Reeves, A., Baker, E., \& Bentley, R. (2017). The impact of persistent poor housing conditions on mental health: A longitudinal population-based study. Preventive medicine, 105, 304-310.

\item Pinto, A. M., Ferreira, F. A., Spahr, R. W., Sunderman, M. A., Govindan, K., \& Meidut\.e-Kavaliauskien\.e, I. (2021). Analyzing blight impacts on urban areas: A multi-criteria approach. Land Use Policy, 108, 105661.

\item Pinto, B. M., Ferreira, F. A., Spahr, R. W., Sunderman, M. A., \& Pereira, L. F. (2023). Analyzing causes of urban blight using cognitive mapping and DEMATEL. Annals of operations research, 325(2), 1083-1110.

\item Ruggiero, R., Rivera, J., \& Cooney, P. (2020). A decent home: The status of home repair in Detroit. Ann Arbor: Poverty Solutions University of Michigan.

\item Sagi, O., \& Rokach, L. (2018). Ensemble learning: A survey. Wiley interdisciplinary reviews: data mining and knowledge discovery, 8(4), e1249.

\item Team, G., Kamath, A., Ferret, J., Pathak, S., Vieillard, N., Merhej, R., ... \& Iqbal, S. (2025). Gemma 3 technical report. arXiv preprint arXiv:2503.19786.

\item Ting, K. M., \& Witten, I. H. (1999). Issues in stacked generalization. Journal of artificial intelligence research, 10, 271-289.

\item Trevethan, R. (2017). Sensitivity, specificity, and predictive values: foundations, pliabilities, and pitfalls in research and practice. Frontiers in public health, 5, 307.

\item US Bureau of the Census (2018). American Community Survey 2018 1-year estimates [Dataset and codebook]. Retrieved from https://censusreporter.org/profiles/16000US2622000-detroit-mi/

\item U.S. Congress. (2021). Exploring How Community Development Financial Institutions Support Underserved Communities: Hearing Before the Committee on Banking, Housing, and Urban Affairs, United States Senate, One Hundred Seventeenth Congress, First Session. U.S. Senate Committee on Banking, Housing, and Urban Affairs.

\item Verma, D., Mumm, O., \& Carlow, V. M. (2023). Generative agents in the streets: Exploring the use of large language models (LLMs) in collecting urban perceptions. arXiv preprint arXiv:2312.13126. https://arxiv.org/abs/2312.13126

\item Walker K, Herman M (2025). tidycensus: Load US Census Boundary and Attribute Data as 'tidyverse' and 'sf'-Ready Data Frames. R package version 1.7.1, https://walker-data.com/tidycensus/.

\item Wang, Z., Majumdar, A., \& Rajagopal, R. (2023). Geospatial mapping of distribution grid with machine learning and publicly-accessible multi-modal data. Nature Communications, 14(1), 5006.

\item Wei, J., Wang, X., Schuurmans, D., Bosma, M., Xia, F., Chi, E., ... \& Zhou, D. (2022). Chain-of-thought prompting elicits reasoning in large language models. Advances in neural information processing systems, 35, 24824-24837.

\item Wolpert, D. H. (1992). Stacked generalization. Neural networks, 5(2), 241-259.

\item Wolpert, D. H., \& Macready, W. G. (2002). No free lunch theorems for optimization. IEEE transactions on evolutionary computation, 1(1), 67-82.

\item Yan, Y., Wen, H., Zhong, S., Chen, W., Chen, H., Wen, Q., Zimmermann, R., \& Liang, Y. (2024). UrbanCLIP: Learning text-enhanced urban region profiling with contrastive language-image pretraining from the web. In Proceedings of the ACM Web Conference 2024 (pp. 4006--4017). Association for Computing Machinery. https://doi.org/10.1145/3589334.3645378

\item Yan, Y., Zeng, Q., Zheng, Z., Yuan, J., Feng, J., Zhang, J., Xu, F., \& Li, Y. (2024). OpenCity: A scalable platform to simulate urban activities with massive LLM agents. arXiv preprint arXiv:2410.21286. https://arxiv.org/abs/2410.21286

\item Yang, Y., \& Liu, X. (1999, August). A re-examination of text categorization methods. In Proceedings of the 22nd annual international ACM SIGIR conference on Research and development in information retrieval (pp. 42-49).

\item Yue, X., Wang, Y., Zhao, Y., \& Zhang, H. (2022). Estimation of urban housing vacancy based on daytime housing exterior images--a case study of Guangzhou in China. ISPRS International Journal of Geo-Information, 11(6), 349.

\item Zeng, T., Wu, L., Peduto, D., Glade, T., Hayakawa, Y. S., \& Yin, K. (2023). Ensemble learning framework for landslide susceptibility mapping: Different basic classifier and ensemble strategy. Geoscience Frontiers, 14(6), 101645.

\item Zhang, Y., Liu, J., \& Shen, W. (2022). A review of ensemble learning algorithms used in remote sensing applications. Applied Sciences, 12(17), 8654.

\item Zhou, Z. H. (2025). Ensemble methods: foundations and algorithms. CRC press.

\item Zou, S., \& Wang, L. (2022). Mapping individual abandoned houses across cities by integrating VHR remote sensing and street view imagery. International Journal of Applied Earth Observation and Geoinformation, 113, 103018.
\end{enumerate}

\end{document}